\documentclass[letterpaper, 10 pt, conference]{ieeeconf}

\IEEEoverridecommandlockouts
\usepackage{url}
\usepackage{balance}
\usepackage{amsmath}
\usepackage{amssymb,bm}
\usepackage{amsfonts}
\usepackage{enumerate}
\usepackage{tabulary}
\usepackage{multirow}
\usepackage{cite}
\usepackage{lipsum}  
\usepackage{graphicx}
\usepackage{epstopdf}
\usepackage[misc]{ifsym}
\usepackage{color}
\usepackage{xcolor}
\usepackage{xxcolor}
\usepackage{pgf}
\usepackage{pgfplots}
\usepackage{tikz}
\usepackage{float}
\usepackage{ifthen}
\usepackage{forloop}
\usepackage{listings}
\usepackage{lipsum}
\usepackage{booktabs}
\usepackage[lined,boxed,commentsnumbered,linesnumbered]{algorithm2e}
\usepackage{verbatim}
\usepackage{hyperref}

\usepgfplotslibrary{groupplots,fillbetween}
\pgfplotsset{compat=newest}

\usetikzlibrary{arrows,shadows,petri,automata,shapes.geometric}
\usetikzlibrary{positioning,fit,calc}
\usetikzlibrary{arrows.meta}
\usetikzlibrary{spy}
\SetAlCapSkip{0.5em}

\title{\LARGE \bf
Learning-based Bias Correction for Ultra-wideband Localization\\ of Resource-constrained Mobile Robots
}

\author{Wenda Zhao, Abhishek Goudar, Jacopo Panerati, and Angela P. Schoellig
\thanks{The authors are with the \href{http://www.dynsyslab.org}{Dynamic Systems Lab}, Institute for Aerospace Studies, University of Toronto, Canada, and affiliated with the Vector Institute for Artificial Intelligence in Toronto. 
E-mails:
        {\tt \{name.lastname\}@utoronto.ca}}}

\begin{document}
\maketitle
\thispagestyle{empty}
\pagestyle{empty}

\begin{abstract}
Accurate indoor localization is a crucial enabling technology for many robotics applications, from warehouse management to monitoring tasks.
Ultra-wideband (UWB) ranging is a promising solution which is low-cost, lightweight, and computationally inexpensive compared to alternative state-of-the-art approaches such as simultaneous localization and mapping, making it especially suited for resource-constrained aerial robots. Many commercially-available ultra-wideband radios, however, provide inaccurate, biased range measurements.
In this article, we propose a bias correction framework compatible with both \emph{two-way ranging} and \emph{time difference of arrival} ultra-wideband localization.
  Our method comprises of two steps: (i) statistical outlier rejection and (ii) a learning-based bias correction. This approach is scalable and frugal enough to be deployed 
  on-board a nano-quadcopter's microcontroller.
  Previous research mostly focused on two-way ranging bias correction and has not been implemented in closed-loop nor using resource-constrained robots.
Experimental results show that, using our approach, the localization error is reduced by $\sim$18.5\% and 48\% (for TWR and TDoA, respectively), and a quadcopter can accurately track trajectories with position information from UWB only. \end{abstract}

\section{Introduction and Related Work}
\label{sec:intro}

\begin{figure}[!t]
	\includegraphics[]{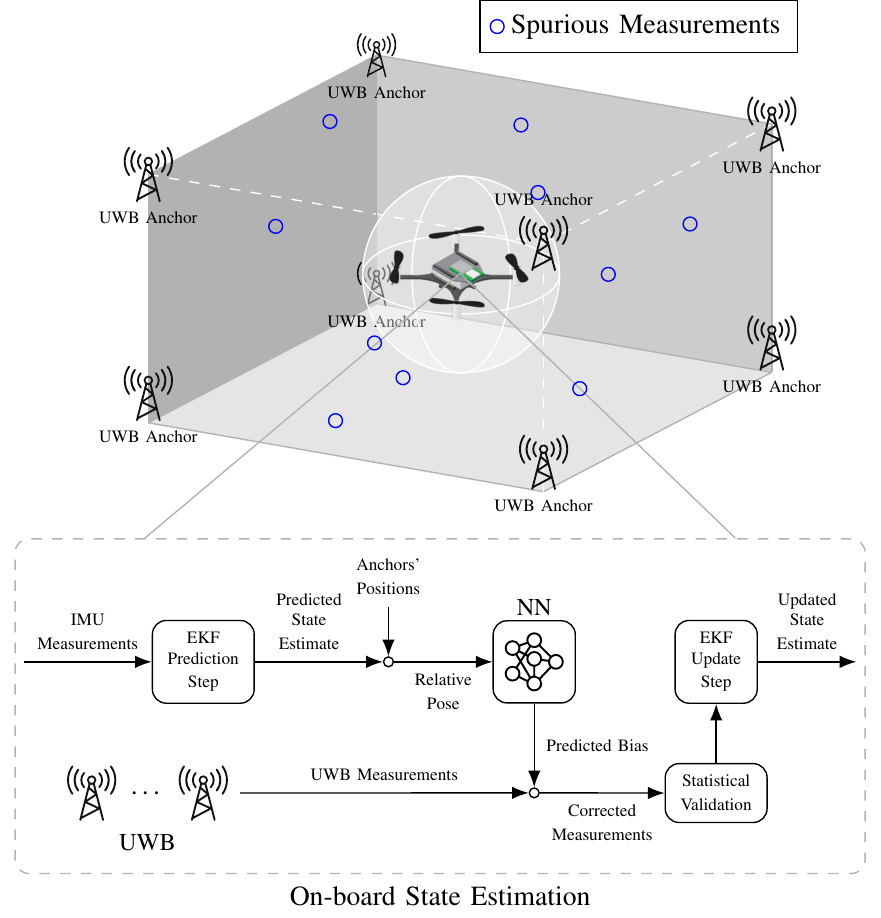}
	\caption{
A stylized depiction of our flight arena and nano-quadcopter platform (top) and the overall schematics of the proposed estimation framework---including the neural network inference and statistical validation steps (bottom). Footage of our work is available at \texttt{\url{http://tiny.cc/uwbBias-Comp}}.
	}
	\label{fig:01}
\end{figure}

Over the last few decades, global navigation satellite systems (GNSS) have become an integral part of our daily lives, providing localization, under open sky, with sub-meter {user range error} anywhere on Earth~\cite{gps}.
Today, indoor positioning systems promise similar radical changes in a plethora of indoor robotics applications (e.g., in warehouses, malls, airports, underground stations, etc.).
Ultra-wideband (UWB) localization technology---based on the same multi-anchor ranging paradigm as GNSSs (see Figure~\ref{fig:01})---in particular, has been shown to provide robust, high-resolution, and obstacle-penetrating ranging measurements~\cite{gezici2005performance, dardari2009ranging}.
UWB chips are already featured in the latest generation of smartphones~\cite{iphone} and, in the near future, are expected to support faster data-transfer as well as accurate, centimeter-level indoor positioning, even  in  cluttered  environments.

Two-way ranging (TWR) and time difference of arrival (TDoA)---based on the time of flight (ToF) of a signal and 
the difference of arrival of multiple, synchronized signals, respectively---are the two main localization approaches that can be implemented using UWB radios to enable autonomous indoor robotics~\cite{hausman2016self, hamer2018self}.
TWR is simpler but less-scalable and more energy-demanding than TDoA.

Nonetheless, many factors can affect the accuracy of UWB measurements, for example, non-line-of-sight (NLOS) communication and multi-path radio propagation can lead to erroneous, spurious measurements. 
Even line-of-sight (LOS) UWB measurements can still be corrupted by spatially-varying UWB biases---those due to the relative pose between UWB radios and their antenna radiation patterns---and systematic biases introduced by multi-path overlays in the UWB pulse~\cite{shen2008effect}. 
The ability to effectively model all these measurement errors is essential to correct them and  guarantee reliable and accurate UWB localization performance.

In this work, we propose a lightweight, two-step measurement correction method---comprising of (i) bias compensation and (ii) outlier rejection---to improve the performance of both TWR and TDoA-based UWB localization.
To achieve (i), we leverage the non-linear representation power of deep neural networks (DNNs). We show that neural networks (NNs) can effectively learn to approximate the spatially varying bias of TWR and TDoA measurements---leading to improved UWB range estimates. 
Then, without the need for raw UWB waveforms~\cite{wymeersch2012machine}, we implement (ii) by rejecting spurious measurements through model-based filtering and statistical validation testing. 
We finally deploy and evaluate our proposed approach on-board a resource-constrained nano-quadcopter.

Multiple approaches have been proposed for the mitigation of UWB TWR measurement errors, yet most of them leverage probabilistic methods to model both the systematic biases and large erroneous measurements caused by NLOS conditions or multi-path propagation.
In~\cite{gururaj2017real}, a channel impulse response (CIR-)based approach detects NLOS propagation from the received UWB waveforms, without the need for prior knowledge of the environment.
In \cite{haggenmiller2019non}, an optimization-based method is used to estimate the {antenna delays} caused by NLOS ranging. 
In~\cite{ledergerber2017ultra} and~\cite{ledergerber2018calibrating}, the authors model and correct UWB pose-dependent biases using sparse pseudo-input Gaussian processes (SPGP) and demonstrate their approach in a quadcopter platform equipped with a Snapdragon Flight computer. While SPGP are already less computationally expensive than standard Gaussian processes, they can become intractable when working with computational resource-constrained nano-quadcopters. 
Furthermore, \cite{ledergerber2017ultra,ledergerber2018calibrating} do not use their proposed bias correction to close the estimation and control loop, as we do in this work.

Fewer research works tackle the modelling of measurement biases when using TDoA rather than TWR.
Pioneering research was conducted in~\cite{prorok2012online,prorok2014accurate}, where an online expectation maximization (EM) algorithm addresses TDoA NLOS measurement errors. 
In~\cite{su2017semidefinite}, a semi-definite programming method is applied to the same problem.
Yet, much of the research on UWB TDoA localization has been conducted in 2D scenario and demonstrated using ground robots.
To the best of our knowledge, our work is the first to enhance UWB TDoA localization through bias correction in 3D scenarios and to use it for closed-loop control of a nano-quadcopter.

Our main contributions can be summarized as follows:
\begin{enumerate}
    \item We present a two-step UWB measurement correction framework comprising of (i) neural network-based bias correction and (ii) statistical outlier detection.
\item We demonstrate the effectiveness of our proposal for both TWR and TDoA-based UWB measurements.
\item We implement the proposed approach onto a resource constrained nano-quadcopter (without the need for custom-designed computing hardware). We show that our method allows for real-time execution during flight and it yields enhanced localization performance for autonomous trajectory tracking. 
\end{enumerate}
The proposed outlier rejection mechanism allows to achieve smooth takeoffs and landings, which would have not been possible using the raw UWB measurements.
Furthermore, neural network-based bias compensation provides a $18.53\%$ and $48.12\%$ reduction in the root mean square error (RMSE) of the position estimation, bringing it to $\sim$20cm for TWR as well as TDoA-based localization. 
To the best of our knowledge, this work is a first demonstration of closed-loop control using a general calibration framework for TWR- and TDoA-based UWB localization, deployed on-board a nano-quadcopter.

 \section{Background}
\label{sec:problem}

UWB measurement errors often have multiple concurrent causes: for example, antenna designs, clock drift, hardware latency, or multi-path propagation. Systematic bias due to UWB antenna radiation patterns and multi-path scenarios due to non-line of sight propagation are the two main phenomena motivating the analysis (and two-step approach) in our work---{yet, we observe that the data-driven nature of our approach (based on machine learning and statistical testing) makes it agnostic to origin of the measurement errors it corrects}.

\subsection{Influence of UWB Antenna Radiation Patterns}

\begin{figure}[!t]
	\includegraphics[]{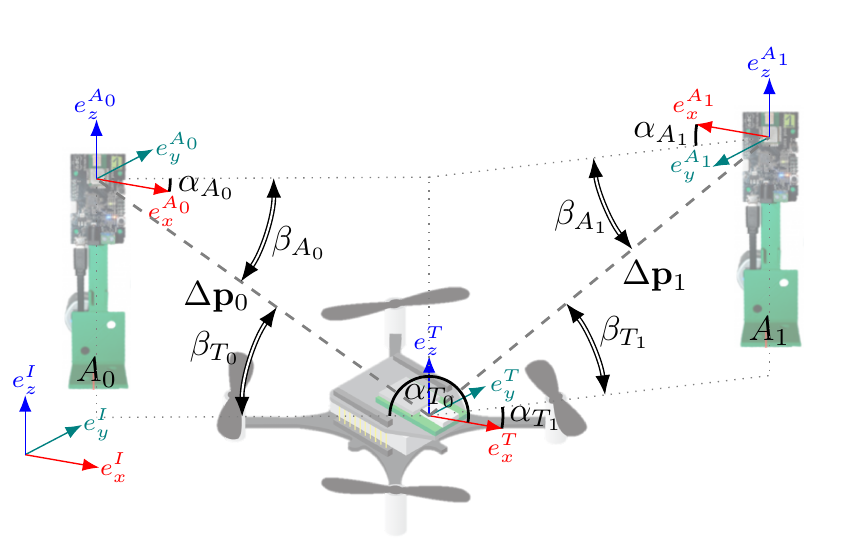}
	\caption{
Schematics of the ranges ($\Delta \mathbf{p}$'s), azimuth ($\alpha$'s) and elevation angles ($\beta$'s) defining the relative poses of tag $T$ and anchors $A_0$, $A_1$ when collecting the systematic bias measurements presented in Figure~\ref{fig:05}. 
	}
	\label{fig:02}
\end{figure}

Measurement errors from off-the-shelf UWB modules can be the result of reduced received signal power~\cite{barral2016assessment,sidorenko2019decawave}. The radiation pattern of doughnut-shaped antennas causes the relative orientation of anchors and tags to have a noticeable impact on the received signal power~\cite{chen2018impact}.
To empirically demonstrate the systematic measurement errors resulting from varying the relative pose between anchors and tags, we placed two DWM1000 UWB anchors at a distance of 4m and collected both TWR and TDoA UWB range measurements for the UWB tag mounted on top of a Crazyflie nano-quadcopter spinning around its own z-axis.
The angles describing the relative poses of anchors and tags are presented in Figure~\ref{fig:02}. 
The TWR and TDoA measurement errors with respect to the bearing of the drone, $\alpha_{T_0}$, are reported in Figure~\ref{fig:05}---TWR measurements only rely on information from $T_0$.
We observe similar trends in the pose-dependent measurement error for both TWR and TDoA, with the latter presenting larger variance.
Figure~\ref{fig:05} also demonstrates that measurement biases strongly depend on the relative pose between UWB radios, and a mean bias correction would fail to capture this relationship.
An in-depth discussion of UWB antenna radiation patterns is beyond the scope of this paper and readers are referred to~\cite{rahayu2012radiation, uvarov2019fundamental, saeidi2019ultra} for further discussions.

\subsection{Influence of NLOS and Multi-path Propagation} 

Multi-path radio propagation is the result of the reflection of radio waves. In indoor scenarios, metal structures, walls, and obstacles are the major causes of multi-path propagation. 
NLOS propagation also often occurs because of the obstacle-penetrating capability of UWB radios. 
NLOS and multi-path propagation results in erroneous signal arrival times, which can lead to significant errors in both TWR and TDoA UWB measurements.
Finally, UWB measurements may also be corrupted by the overlaid UWB pulse caused by multi-path propagation, leading to additional measurement errors~\cite{shen2008effect}.

 \section{Methodology}
\label{sec:methods}

In TWR-based UWB localization systems, the UWB module on a target device (e.g., a mobile robot)---also called \emph{tag}---ranges to a fixed UWB radio---also called \emph{anchor}---to compute its distance based on the ToF of the signal. In TDoA-based localization, {tags} receive UWB radio signals from time-synchronized {anchors}. 
Unlike TWR, TDoA does not require two-way communication, enabling the localization of a larger number of tags. 
However, the performance of TDoA is remarkably more sensitive to noise.

\subsection{Measurement Error Models}
\label{sec:uwb1}
 For a TWR-based localization system with $m$ UWB anchors, if the position of anchor $i$ is denoted as $\mathbf{p}_i = \left[x_i,y_i,z_i\right]^T\in \mathbb{R}^3, i = 1,\dots,m$, and the position of the tag is denoted as $\mathbf{p} = \left[x, y, z\right]^T\in \mathbb{R}^3$, then, the TWR measurement model for anchor $i$ can be written as:
\begin{equation}
\label{Eq: twr measurement model}
    \tilde{r}_{i} = r_{i} + b_i + \epsilon_{i},
\end{equation}
where $r_{i} = \|\mathbf{p}-\mathbf{p}_i\|$ is the true distance between anchor $i$ and the tag, $b_i$ is the bias term that we want to model, $\epsilon_{i} \sim \mathcal N (0 , \sigma_i^2)$ is a measurement noise with the variance $\sigma_i^2$, and $\|\cdot\|$ is the Euclidean norm. 
\label{sec:uwb2}
TDoA-based localization computes position estimates using the arrival time difference of signals from multiple anchors. The TDoA error model for any two anchors $i$ and $j$ can be written as:
\begin{equation}
\label{Eq: tdoa measurement model}
    \tilde{r}_{ij} = r_{ij} + b_{ij} + \epsilon_{ij},
\end{equation}
where $r_{ij} = \|\mathbf{p}-\mathbf{p}_i\| - \|\mathbf{p}-\mathbf{p}_j\|$ is {the difference in range measurements with respect to anchors $i$ and $j$}, $b_{ij}$ is the TDoA measurement bias {relative to anchors $i$ and $j$}, and $\epsilon_{i,j} \sim \mathcal N (0 , \sigma_{ij}^2)$ is the TDoA measurement noise with the variance $\sigma_{ij}^2$.

\subsection{Neural Network Bias Correction}
\label{sec:learning}

To model---and predict, in real-time, during flight---the pose-dependent measurement bias, we leverage the fast inference capabilities of a pre-trained neural network.
Since multi-path and NLOS propagation effects depend on a particular indoor environment, we only use the NN to explicitly model the pose-dependent bias.
Neural networks are highly flexible in the number of inputs they take and scalable with the amount training data that they can exploit, making them a suitable tool to capture the complex relationship between UWB measurement biases and relative poses of the tag and anchors.

Given sufficient network capacity and training data, the UWB measurement bias can be described by a nonlinear function $b = f(\mathbf{x})$ captured by a NN with input feature vector $\mathbf{x}$. 
Thus, the UWB measurement \emph{after NN bias compensation}, $r^{\star}$, can be computed as:
\begin{equation}
    \label{Eq:DNN bias compensation}
    r^{\star} = \tilde{r} - f(\mathbf{x}) = r  + \epsilon,
\end{equation}
where $r$, $\tilde{r}$ and $\epsilon$ were defined for TWR and TDoA in \eqref{Eq: twr measurement model} and \eqref{Eq: tdoa measurement model}, respectively.
For TWR, we define  $\mathbf{x}$ as $\left[\Delta \mathbf{p}_i^T, \bm{\theta}^T \right]^T$, where $\Delta \mathbf{p}_i = \left[x_i - x, y_i - y, z_i - z\right]^T$ represents the relative distance between anchor $i$ and the tag and $\bm{\theta} = \left[\phi, \theta, \psi\right]^T$ is the roll, pitch, and yaw of the quadcopter. 
For TDoA, we define $\mathbf{x}$ as $\left[\Delta \mathbf{p}_i^T, \Delta \mathbf{p}_j^T, \bm{\theta}^T \right]^T$ in which $\Delta \mathbf{p}_i$ and $\Delta \mathbf{p}_j$ represent the relative distances between the tag and anchor $i$ and $j$, respectively. 
As we used fixed anchors, we do not include their poses in our feature vectors~\cite{ledergerber2017ultra}. Studying our approach's ability to generalize to different anchor constellations by doing so is left as future work.
{ }

\subsection{Outlier Rejection}
\label{sec:outlier}

\begin{figure}[!t]
	\includegraphics[]{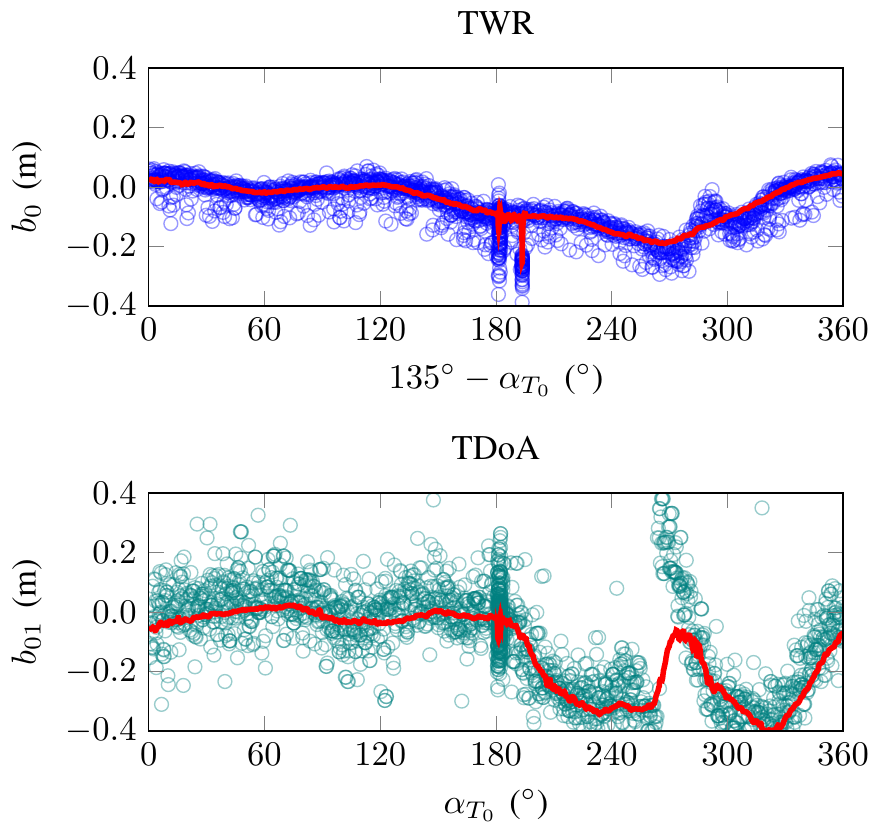}
	\caption{
The neural network's inferred bias (in red) with respect to the tag's varying azimuth angle towards anchor $T_0$, $\alpha_{T_0}$, plotted against the UWB raw measurements.
	For TWR, $\alpha_{A_0}$ is $-45^\circ$, $\beta_{A_0}$ is $20^\circ$, and $||\mathbf{r}_{0}$ is 1.5m; for TDoA, $\alpha_{A_{0,1}}$ are $0^\circ$, $\beta_{A_{0,1}}$ are $14^\circ$, and $||\mathbf{r}_{0,1}||$ are 2m.
	}
	\label{fig:05}
\end{figure}

\begin{figure}[!t]
	\includegraphics[]{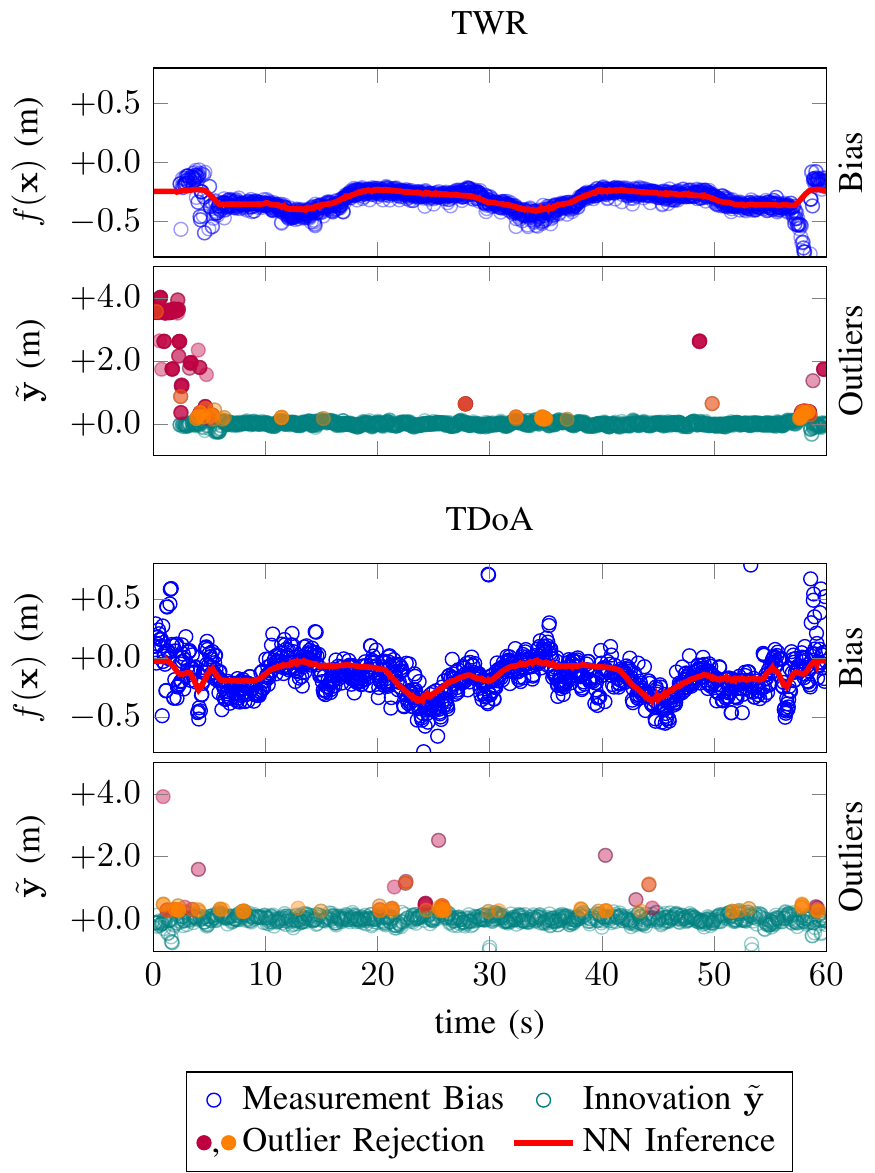}
	\caption{
UWB measurement bias $f(\mathbf{x})$ prediction performance of the trained neural network (in red) compared to the actual measurement errors (blue dots) as well as the role of model-based filtering (purple dots) and statistical validation (orange dots) in rejecting outlier measurement innovations $\tilde{y}$ (teal dots) during a 60'' flight experiment.
	}
	\label{fig:04}
\end{figure}

Having applied the NN-based bias correction presented in Subsection~\ref{sec:learning}, we first use a model of a robot's dynamics---in our case, a quadcopter---to filter inconsistent UWB range measurements. 
Given the estimated velocity $\mathbf{v}$ and maximum acceleration $\mathbf{a}_{max}$---the maximum distance ${d}_{max}$ a quadcopter can cover during time $\Delta t$ is:
\begin{equation}
\label{Eq: maximum distance}
    {d}_{max} = \|\mathbf{v}\Delta t+\frac{1}{2}\mathbf{a}_{max}\Delta t^2\|.
\end{equation}
Let $\tilde{\mathbf{y}}_k$ be the measurement innovation in the extended Kalman filter (EKF) at time step $k$,
we reject unattainable measurements 
before fusing them into the EKF. 
For TWR, we discard UWB measurements whose innovation term is larger than  $d_{max}$. For TDoA, the UWB measurements are rejected if the innovation term is larger than $2d_{max}$. 

Then, we use a statistical hypothesis test inspired by~\cite{hausman2016self} to further classify potential outlier measurements. In the standard EKF, the covariance of the measurement innovation is expressed as:
\begin{equation}
\label{Eq: convariance of measurement innovation}
    \mathbf{S}_k = \mathbf{G}_k\check{\mathbf{P}}\mathbf{G}_k^T + \mathbf{R}_k,
\end{equation}
where $\mathbf{G}_k = \frac{\partial g(\check{\mathbf{x}}) }{\partial \check{\mathbf{x}}}$ is the Jacobian of the measurement model, $\check{\mathbf{P}}$ is the prior covariance of the estimated states, and $\mathbf{R}_k$ is the covariance of measurement $\mathbf{y}_k$. Since the measurement innovation vector $\tilde{\mathbf{y}}$ is assumed to be distributed according to a multivariate Gaussian with covariance matrix $\mathbf{S}$, the normalized sum of squares of its values should be distributed according to the $\chi^2$ distribution. We use the $\chi^2$ hypothesis test to determine whether a measurement innovation is likely coming from this distribution:
\begin{equation}
\label{Eq: chi-squared test}
    \tilde{\mathbf{y}}_k^T\mathbf{S}_k^{-1}\tilde{\mathbf{y}}_k \leq \chi^2(0.95).
\end{equation}
Values above this threshold are labeled as outliers and not carried on into the EKF. 
The outlier rejection results, for both TWR and TDoA, are shown in the second and fourth plot of Figure~\ref{fig:04}. 
The outliers detected by different methods are labeled with different colors: the purple dots indicate outliers labeled by the model-based approach while orange dots are those detected through the $\chi^2$ test.

 \section{Experimental Results}
\label{sec:results}

We use Bitcraze's Crazyflie 2.0 nano-quadcopter and the Loco Positioning System (LPS)'s UWB DW1000 modules as our research platforms. 
Our calibration approach and a non-linear controller~\cite{mellinger2011minimum} run on the Crazyflie STM32 microcontroller within the FreeRTOS real-time operating system\footnote{\texttt{\url{https://www.freertos.org/}}}. 
We equipped a cuboid flying arena ($7$m $\times$ $8$m $\times$ $3$m) with 8 UWB anchors, one for each vertex (see Figure~\ref{fig:06}). 
The anchor positions were measured using a Leica total station theodolite.
For all experiments, the ground truth position of the Crazyflie drone was provided by 10 Vicon cameras.
In considerations of the microcontroller computational constraints, for our NN, we choose a feed-forward architecture comprising of 50 neurons with rectified linear activation functions (ReLUs) in each of two layers.
The network was trained using \texttt{PyTorch}~\cite{paszke2017automatic}.
To port it to the Crazyflie's microcontroller, we re-use \texttt{PyTorch}'s trained weights in a plain C re-implementation.
Since the DW1000 modules in the LPS provide UWB measurements every $5$ms, the NN inference rate runs at $200$Hz during flight as well.  
Our outlier rejection method is also implemented in plain C and merged into the on-board EKF.

\subsection{Data Collection and Training}
\label{sec:data}

\begin{figure}[!t]
	\includegraphics[]{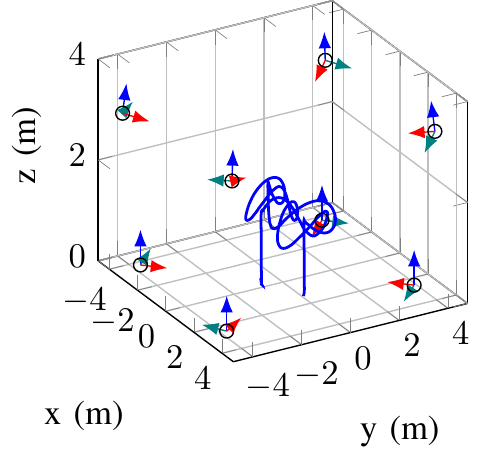}
	\includegraphics[]{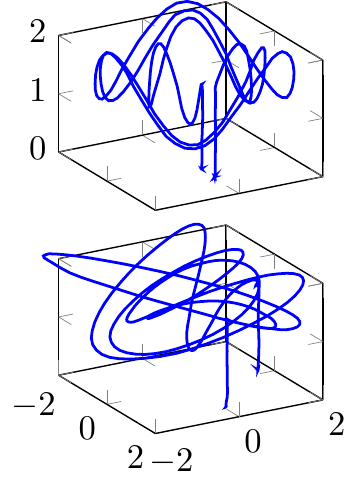}
	\caption{Left: three-dimensional plot of our flight arena showing the positions and poses of the eight UWB DW1000 anchors (each facing towards its own x-axis, i.e., the red versor). Right: two of the training trajectories we flew to collect the samples that we used to train our neural network-based bias estimator.
	}
	\label{fig:06}
\end{figure}

To train our NN, we collected UWB measurements through tens of real-world flights along generic trajectories of a Crazyflie equipped with a low-cost IMU, UWB tag, and reflective motion capture markers.
A few example training trajectories are shown in Figure~\ref{fig:06}. 
Over $700'000$ UWB measurements were logged at $50$Hz, compared against the motion capture information to compute the corresponding range error, and added to our training set. 
Note that, to separate the bias correction problem from outlier rejection, we exclusively trained our NN with measurement whose 
actual bias less within a threshold $\Xi$ of 0.7m.
We used $90\%$ of this dataset for training and the rest for hyper-parameter tuning.
As an optimizer, we chose mini-batch gradient descent\cite{robbins1951stochastic}. 

\subsection{Flight Experiments}
\label{sec:experiments}

We demonstrate the estimation and closed-loop performance of the methods presented in Section~\ref{sec:methods} by flying a Crazyflie quadcopter along circular trajectories in the x-y plane (at varying altitudes) which were not part of the trajectories used for training.
Footage of both TWR and TDoA UWB-based flights is available online\footnote{\texttt{\url{http://tiny.cc/uwbBias-Comp}}}.

\begin{figure}[!t]
	\includegraphics[]{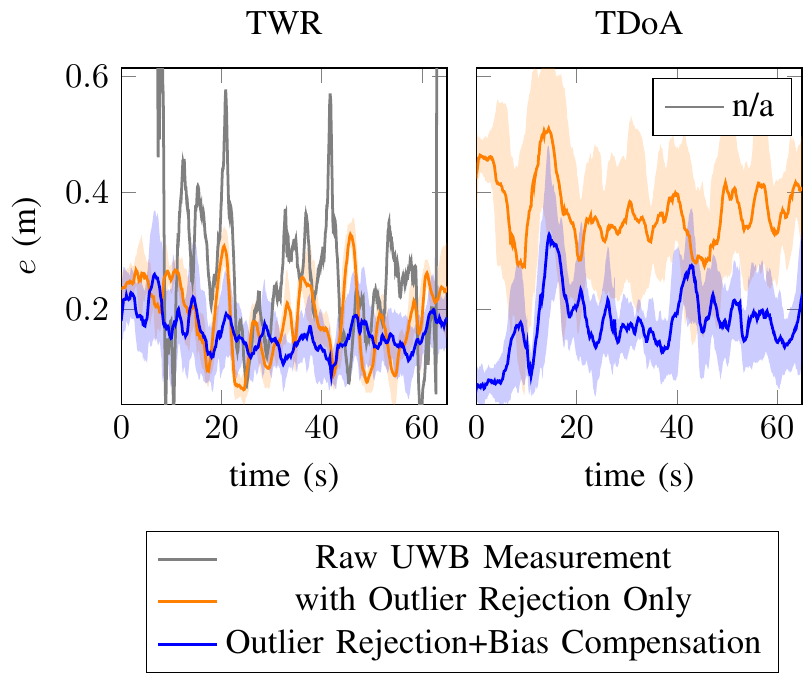}
	\caption{
Estimation performance $e$ with outlier rejection only (in orange) and with the proposed outlier rejection and neural network calibration step (in blue) using both UWB ranging modes (TWR and TDoA). 
    Results are averaged over 10 experiments, standard deviations are presented as shadowed areas.
    The raw UWB measurement error is reported for TWR but not TDoA as the latter did not allow for reliable take-off.
	}
	\label{fig:07}
\end{figure}

\subsubsection{UWB Localization Estimation Performance}
without the proposed bias compensation and outlier rejection strategies, TWR-based measurements show large estimation errors
when the quadcopter is on the ground. This is likely due to more severe reflections and multi-path effects. Such large estimation errors make it difficult for the quadcopter to takeoff reliably. These adverse effects are even more severe for TDoA-based localization, leading to repeated take-off failures.
Consequently, we did not collect in-flight results for raw TDoA measurements.
The NN's fit of the UWB measurement biases---for both TWR and TDoA---is shown in Figure~\ref{fig:05}.
Similar results---but for a $60''$ flight trajectory rather than a full revolution around the Crazyflie's own z-axis---are presented in the first and third plot of Figure~\ref{fig:04}.

A comparison between the estimation error of (i) the UWB raw localization estimate, (ii) the estimate enhanced with outlier rejections, and (iii) the estimate enhanced with both outlier rejection and NN-based bias compensation is shown in Figure~\ref{fig:07} for both TWR and TDoA measurements.

\begin{figure}[!t]
	\includegraphics[]{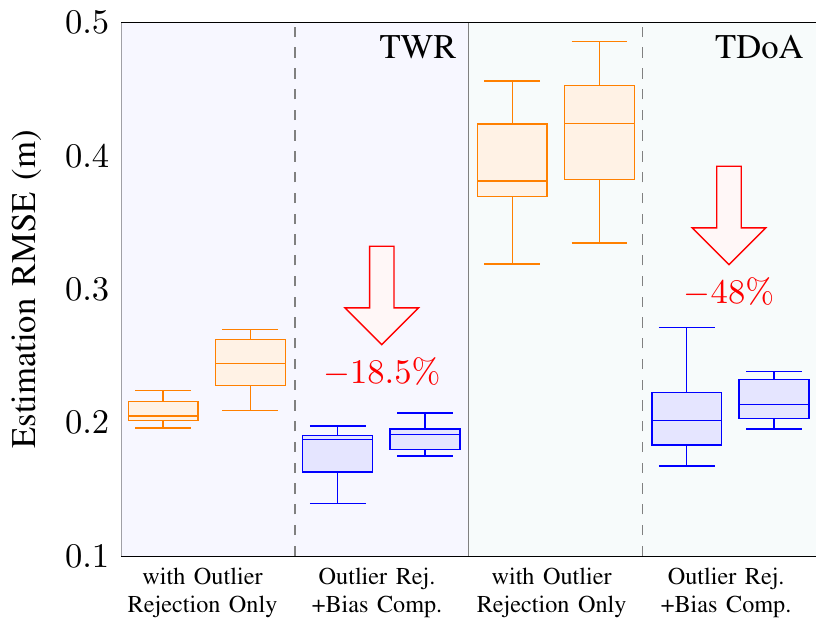}
	\caption{
Root mean square error (RMSE) of the quadcopter position estimate before (in orange) and after (in blue) the neural network calibration step for both TWR and TDoA ranging modes. Each pair of box plots refers to a planar reference trajectory (left of each pair) and a reference trajectory with varying $z$ (right of each pair), showing a greater performance enhancement for the latter.
	All box plots represent distributions over 10 experiments. 
}
	\label{fig:09}
\end{figure}

{We compute the estimation error $e=\|\mathbf{p}_v-\mathbf{p}\|$ as the offset from the ground truth position $\mathbf{p}_v = \left[x_v, y_v, z_v\right]^T\in \mathbb{R}^3$ provided by the Vicon system.}
The distribution of the root-mean-square (RMS) estimation error $e$---when using (i) outlier rejection only or (ii) both bias compensation and outlier rejection---is given in Figure~\ref{fig:09}, for both TWR and TDoA. 
As expected, TWR-based ranging results in better localization performance than TDoA. 
However, we observe that, with our NN-based bias compensation, the average RMS error of TDoA localization is around $0.21$m, which is comparable to that of TWR-based localization ($\sim$$0.19$m). 
Thanks to DNN-based bias compensation, the average reduction in the RMS error is $\sim$$18.5$\% and $48$\%, for TWR and TDoA, respectively. 
The diminishing returns provided by bias compensation for TWR are a consequence of the already relatively accurate localization achieved by outlier rejection only.
Nonetheless, we observe that NN-based bias correction improves estimation, especially when reference trajectories exhibit a larger variance on the $z$ axis.
Most notably, Figure~\ref{fig:09} suggests that bias compensation might help closing the performance gap between TWR- and TDoA-based localization.

\subsubsection{UWB Localization Performance in Closed-loop}
\begin{figure}[!t]
	\hspace{0.6cm}\includegraphics[]{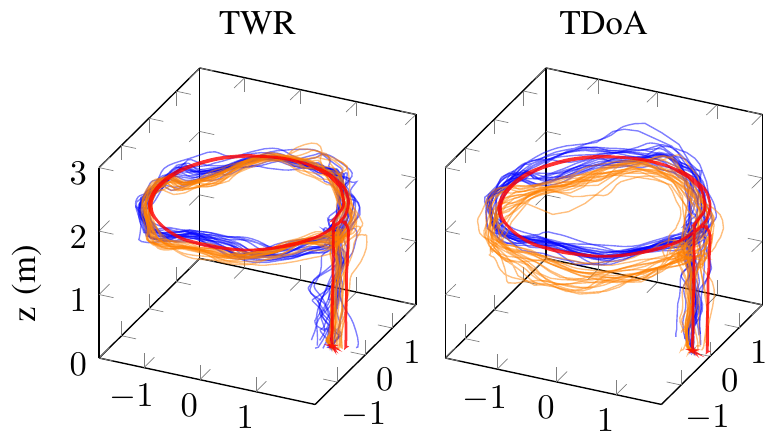}
	\includegraphics[]{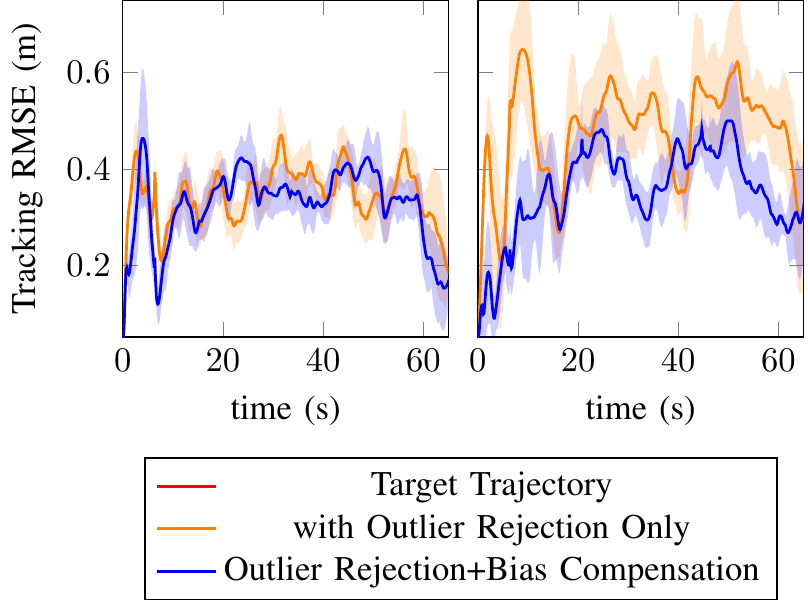}
\caption{
Flight paths and RMS of tracking error $e_t$ with outlier rejection only (in orange) and with the proposed outlier rejection and neural network calibration step (in blue) for a planar reference trajectory using both UWB ranging modes (TWR and TDoA). Results are averaged over 10 experiments, standard deviations are presented as shadowed areas.
}
	\label{fig:08}
\end{figure}

to evaluate the real-world performance of the proposed calibration method, we study the trajectory tracking error along our testing trajectories with a desired velocity of $0.375$m/s.
We repeated all of our experiments (i.e. using the two UWB ranging modes, with outlier rejection only and both outlier rejection and bias compensation), each lasting $\sim$$60-80''$, 10 times.
Closed-loop control with raw UWB localization was not attempted to guarantee flight safety. 
At each time step, we compute the trajectory tracking error $e_t = \|\mathbf{p}_v-\mathbf{p}_c\|$, where $\mathbf{p}_c = \left[x_c, y_c, z_c\right]^T\in \mathbb{R}^3$ is the commanded waypoint.
The results of this test campaign are presented in Figure~\ref{fig:08}.

The quadcopter trajectories during these flight tests are displayed in the top 3D plots. 
The solid red lines show the ideal trajectory flown by a Crazyflie using position information coming from the Vicon motion capture system. 
Orange lines present the flight performance of a Crazyflie using only outlier rejection, while blue lines refer to a Crazyflie exploiting both outlier rejection mechanisms and NN-based bias compensation. 
In the bottom row of Figure~\ref{fig:08}, the RMSE of the tracking error is plotted against the flight time. 
The blue and orange lines show the tracking error---with and without NN-based bias correction, respectively---averaged over 10 flight experiments. Shadowed areas represent the respective standard deviations.
For both TWR- and TDoA-based localization, the position errors along the two testing trajectories are generally reduced with NN bias compensation, especially for TDoA and along the $z$-axis.

\section{Conclusions}
\label{sec:conclusions}

In this article, we presented a two-step methodology to improve UWB localization---for both TWR- and TDoA-based measurements.
We used a lightweight NN to model and compensate for pose-dependent and spatially-varying biases and an outlier rejection mechanism to filter spurious measurements.
We also demonstrated that our proposed approach can be effectively implemented and run on-board a resource constrained nano-quadcopter.
Through several real-world flight experiments tracking different trajectories, we showed that we are able to improve localization accuracy for both TWR and TDoA, granting safer indoor flight.
In our future work, we will include the anchors' pose information to allow our method to further generalize to previously unobserved indoor environments, with different anchor configurations.

\balance

\balance
\end{document}